# A Hybrid SOM and K-means Model for Time Series Energy Consumption Clustering

Farideh Majidi
Department of Computer Engineering
Islamic Azad University, South Tehran Branch
Tehran, Iran
st_f.majidi@azad.ac.ir

*Abstract*— Energy consumption analysis plays a pivotal role in addressing the challenges of sustainability and resource management. This paper introduces a novel approach to effectively cluster monthly energy consumption patterns by integrating two powerful techniques: Self-organizing maps (SOM) and K-means clustering. The proposed method aims to exploit the benefits of both of these algorithms to enhance the accuracy and interpretability of clustering results for a dataset in which finding patterns is difficult. The main focus of this study is on a selection of time series energy consumption data from the "Smart meters in London" dataset. The data was preprocessed and reduced in dimensionality to capture essential temporal patterns while retaining their underlying structures. The SOM algorithm was utilized to extract the central representatives of the consumption patterns for each one of the houses over the course of each month, effectively reducing the dimensionality of the dataset and making it easier for analysis. Subsequently, the obtained SOM centroids were clustered using K-means, a popular centroid-based clustering technique. The experimental results demonstrated a significant silhouette score of 66%, indicating strong intra-cluster cohesion and inter-cluster separation which confirms the effectiveness of the proposed approach in the clustering task.

## I. Introduction

The efficient management of energy resources is an imperative task in the pursuit of sustainability and environmental responsibility. Among the various sectors contributing to energy consumption, residential housing stands as a significant contributor, warranting rigorous investigation into energy usage patterns and behaviors. The advent of smart metering technologies has facilitated the collection of granular energy consumption data [1], offering new opportunities for detailed analysis and insights. The dataset used in this paper was "Smart meters in London" from the Kaggle website, which contains the energy consumption readings of smart meters of 5567 houses in London from 2011 to 2014. The dataset also contains the weather data which was of no use for the purposes of this study. Since the mission of this research was to discover the patterns of monthly energy consumption in different houses and try to put them in groups so that it would benefit load distribution management, the best method to unravel these patterns is clustering. Clustering, a fundamental technique in data analysis, enables the grouping of similar data points, thereby revealing underlying patterns and structures within the dataset [2, 3]. Particularly, clustering energy consumption data from households can aid in identifying distinct consumption profiles, which are pivotal for the design of tailored energy efficiency strategies and load distribution systems in general [4, 5]. Traditional clustering algorithms, however, may struggle to handle the temporal dynamics inherent in time series data, leading to suboptimal results and limited interpretability [6]. It's important to note that only a single clustering algorithm hardly finds any pattern in big and diverse datasets such as energy consumption data collected from smart meters. This paper presents an innovative approach that seamlessly integrates two powerful clustering techniques: Self-organizing maps and K-means. Self-organizing maps are neural network-based unsupervised learning algorithms capable of capturing complex relationships in high-dimensional data [7, 8, 9]. They excel at preserving the topology of the input space while reducing dimensionality, thus enabling the representation of intricate temporal patterns. Additionally, SOM works very well with time series data and is able to find patterns that can be difficult for other clustering methods [10]. K-means, on the other hand, is a centroid-based clustering algorithm that efficiently partitions data into clusters based on similarity to cluster centroids [11, 12]. The combination of these methods aims to leverage the strengths of each algorithm, leading to enhanced clustering performance and a more meaningful interpretation of results [13, 14]. This study focuses on the task of clustering energy consumption patterns from a part of the introduced dataset comprising time series data from 50 residential houses over the first half of a two-year period (2012 and 2013) since the pattern changes drastically in the second half of the year. The principal aim of this study revolves around the extraction and comprehensive analysis of unique consumption profiles. These profiles hold the potential to offer valuable insights for the optimization of targeted load distribution systems in different energy resources [15, 16, 17]. Specifically, this study's approach involves a transformative representation of the data, wherein individual house-level time series data is aggregated into monthly clusters. This innovative technique allows for the identification of previously unrecognized energy consumption patterns across households. By shifting the focus from the conventional analysis of individual house-level time series data – a labor-intensive endeavor due to the inherent irregularities in consumption patterns – to a monthly cluster-based approach, this study offers a more efficient and effective means of analyzing energy consumption patterns. By utilizing SOM to identify central representatives of the temporal patterns and subsequently applying K-means to cluster these representatives, the goal is to provide a comprehensive framework for understanding the diversity of energy consumption behaviors within the sampled houses.

The remainder of this paper is organized as follows: Section II provides a brief overview of related work in the fields of energy consumption analysis and the integration of SOM and K-means. Section III outlines the methodology, detailing the data preprocessing steps, the SOM algorithm, PCA, and the subsequent K-means clustering process. Section IV presents the findings and results, including the evaluation metrics used to assess the clustering performance and their potential applications in energy management strategies. Finally, Section V concludes the paper, by summarizing the contributions and highlighting the significance of the proposed approach in advancing the understanding of energy consumption patterns, and potential future work will be suggested as well.

## II. RELATED WORK

In [18], an innovative clustering tool integrating Self-organizing maps with the k-means algorithm was introduced. This amalgamation offers a robust approach to partitioning data into distinct clusters, considering intricate interdependencies among features. Addressing the intricate task of data clustering, especially when confronted with voluminous databases and intricate feature relationships that may obscure potential clusters, demands the application of sophisticated computational methodologies. SOMs, a powerful technique, excel at discerning patterns and mapping high-dimensional databases onto two-dimensional representations. Yet, in some scenarios, these maps exhibit a level of detail that proves overly intricate for certain decision-making processes. The outcome often is an elaborate representation that may not directly serve decision-making needs, necessitating additional steps. Within this study, a novel strategy emerges, proposing the hybrid utilization of SOMs as a preliminary step. Initially, a SOM is employed to craft a feature map exhibiting an optimal architecture attained through the minimization of an efficiency index. Subsequently, by operating on a SOM-derived dissimilarity matrix, the k-means algorithm undertakes a process of map coarsening. This transformation yields a reduced number of clusters, more aptly suited for diverse decision-making contexts. The effectiveness of this hybrid approach is exemplified through three distinct case studies pertaining to different aspects of Water Distribution Systems (WDSs). This study underscores the potential for enhancing the management, operation, and planning of WDSs through judicious data utilization. Moreover, it highlights the escalating challenge of practical data application as databases expand in size. The positive outcomes witnessed across the three case studies (encompassing planning, operation, and management of WDSs) hint at the versatility of the proposed hybrid approach in addressing challenges across diverse domains, notably in environmental contexts.

In [19], a novel hybrid algorithm combining K-means and SOM neural network techniques is introduced as a potential solution for breast cancer prediction within the medical domain. Through empirical evaluations, the hybrid algorithm showcases its capacity to amalgamate the strengths of the K-means and SOM algorithms, while effectively addressing their individual limitations. By juxtaposing the outcomes against those of the standalone K-means algorithm and traditional SOM algorithm, distinct advantages emerge. The hybrid algorithm attains a superior level of accuracy in comparison to the standalone K-means approach. Similarly, when contrasted with the conventional SOM algorithm, the hybrid approach not only advances accuracy levels but also substantially reduces computational runtimes. Additionally, the scope of the experimental evaluation is bounded by the extent of the available dataset. To solidify the algorithm's efficacy and generalizability, a broader range of experimental datasets should be incorporated into future investigations. This expansion is crucial in ascertaining the algorithm's robustness across diverse scenarios, enhancing its credibility as a predictive tool.

In [20], The fusion of k-means and SOM methodologies has been effectively employed to facilitate regional clustering grounded in air quality parameters within Makassar City. The outcomes of these clustering endeavors furnish valuable insights into the prevailing air pollution levels across distinct geographical areas during both diurnal and nocturnal periods. Notably, a lower air quality index within a given regional cluster is indicative of elevated pollution levels. The task of determining the appropriate number of clusters was addressed by analyzing the Within-Cluster Sum of Squares (WSS) graph. This analysis revealed that during the night, an optimal configuration encompassing 7 regional clusters yields the most meaningful partitioning of data, whereas, in daylight hours, the clustering coalesces into 4 regional clusters. This distinction underscores variations in the dispersion of pollution characteristics, demarcating differences between day and night air pollution profiles. The observed heterogeneity in air pollution distribution across regional clusters can be attributed to diverse factors, including variations in traffic density, industrial activities, and ongoing developmental projects within each locality. The synergy between k-means and SOM has facilitated the categorization of Makassar City's regions based on air quality parameters. This endeavor has revealed nuanced pollution disparities between daytime and nighttime, underlining the influence of various factors.

In [21], the author has contributed to the literature by introducing two distinctive approaches: cluster-wise regression and cluster validation techniques. These novel contributions were aimed at addressing distinct yet interconnected challenges in modeling building energy consumption. Cluster-wise regression emerged as a noteworthy advancement, offering a novel perspective on predictive accuracy. Its application showcased remarkable precision in forecasting energy consumption patterns. However, it is important to note that the stability of the derived clusters exhibited some variability, suggesting room for improvement in this aspect. In comparison, conventional methods such as K-means and model-based clustering have historically been employed in the domain of building energy consumption analysis. These techniques, while demonstrating relatively stable cluster outcomes, have shown limitations in prediction accuracy. Notably, certain clusters identified through these methods yielded considerably poor predictions, an aspect that might not have been immediately discernible during the initial stages of analysis. This research has thus underscored the necessity of a careful method selection process based on specific use cases. The inherent trade-off between the goals of predictive accuracy and cluster stability necessitates a nuanced approach. While cluster-wise regression excels in predictive accuracy but presents challenges in cluster stability, K-means and model-based clustering exhibit differing strengths and weaknesses in these aspects. The selection of an appropriate methodology should be driven by the specific objectives of the analysis, ensuring alignment with the priorities and constraints of the study at hand.

III. METHODOLOGY

In this section, the novel approach that combines Self-organizing maps and K-means algorithms for clustering time series data will be introduced. The proposed method capitalizes on the distinct strengths of both techniques to enhance the accuracy and effectiveness of time series clustering. The initial phase of the hybrid method involves the application of SOM to facilitate a preliminary clustering of the time series data. Prior to initiating the initial clustering stage, data normalization is a prerequisite to facilitate pattern discovery by the model. To fulfill this objective, the MinMax scaler methodology was employed, chosen for its efficacy in achieving this normalization target. The MinMax scaling technique, often employed in data preprocessing for machine learning and clustering tasks, serves the purpose of normalizing data features within a specific range. This approach ensures that all features have comparable magnitudes, thus preventing any undue influence on the clustering process due to the differences in scale among features. The MinMax scaler operates by transforming the original data values of a feature into a new range, which is from 0 to 1 in this paper.

SOM is a neural network-based technique renowned for its ability to map high-dimensional data onto a lower-dimensional grid while preserving their inherent relationships. The SOM algorithm generates a feature map composed of nodes, with each node representing a specific feature vector. During training, the SOM adjusts its nodes iteratively to align with the input data distribution, forming clusters within the map. Once the SOM training converges, the center of each cluster is extracted within the SOM feature map, it's important to note that SOM does a very good job discovering these patterns although it requires the supervision of a human to determine the number of clusters for each month so that there would be no empty or overfilled clusters that harm the output of the first clustering phase. These cluster centers serve as representative vectors encapsulating the characteristics of the data points associated with each cluster for a certain period of time. Extracting these centers is accomplished by identifying the node with the highest activation for each cluster. These extracted centers encapsulate the core features of the data in a reduced-dimensional space and can be a new representation of the data since there is no need to find the patterns for each house separately, but the monthly clustered data and its centers are the new inputs for the next clustering phase which determines which parts of this newly generated data can be put in the same clusters.

Upon the generation of SOM-derived cluster centers, an additional layer of sophistication is introduced by integrating Principal Component Analysis (PCA). PCA is a statistical technique used to reduce the dimensionality of high-dimensional data while retaining as much of its variance as possible. In the context of the proposed methodology, PCA is applied to the cluster centers extracted from SOMs. This process involves transforming the high-dimensional centers into a new, lower-dimensional subspace while preserving the most salient information. Subsequently, the cluster centers, now reduced in dimensionality through PCA, are utilized as inputs for the k-means algorithm, resulting in more homogenous clusters.

K-means is a well-established centroid-based clustering method that partitions data into distinct clusters by iteratively assigning data points to the nearest cluster center. In this hybrid approach, K-means operates on the dimensionally reduced extracted SOM cluster centers, allowing it to focus on a more refined, abstract representation of the original time series data. Once the K-means algorithm converges on the cluster centers, its application is extended to the entire dataset. Each data point is assigned to the cluster associated with the nearest cluster center determined during the K-means phase. This results in a comprehensive clustering of the complete time series dataset based on the insights gained from the SOM-derived cluster centers of energy consumption data. The flowchart of the introduced model is shown in Figure 1.

To accomplish this, a two-step clustering method was employed. The first step involved a supervised approach to determine the optimal number of clusters for monthly data segmentation. This step was particularly important to ensure that our subsequent clustering process would yield meaningful and interpretable results. After careful consideration, it was determined that using 88 clusters would be the most suitable choice for this phase of the analysis. To further streamline the data and enhance its efficiency for clustering, Principal Component Analysis was applied to reduce the dimensionality of the cluster centers. This step allowed the model to retain the essential features of the data while reducing computational complexity. Following dimensionality reduction, the K-means algorithm was employed to perform the actual clustering of the data points based on the reduced cluster centers. In the second clustering phase, determining the optimal number of clusters was crucial. To make this decision, the silhouette score was used, which is a metric widely employed to evaluate the quality of clusters. The silhouette score aided in selecting the most appropriate number of clusters to effectively represent and segment the energy consumption data.

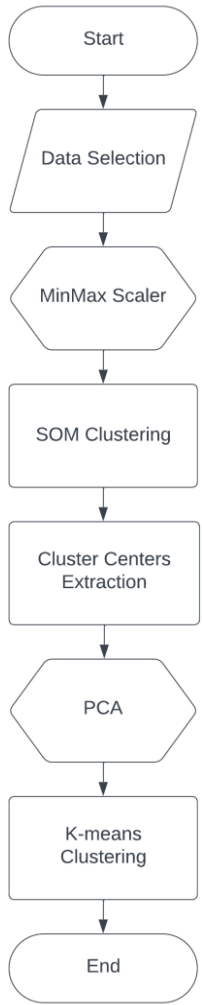

Figure 1.  Flowchart of the SOM and K-means fusion model

This multi-step approach made it possible to systematically organize and analyze the energy consumption patterns within the dataset, ensuring that this clustering process was both well-informed and data-driven. The ultimate aim of this research was to extract meaningful insights from these clusters, providing a foundation for more informed decision-making and predictions in the domain of energy consumption analysis.

## IV. Results and Discussion

In this section, the significant findings stemming from the application of the integrated SOM-k-means methodology for uncovering energy consumption patterns within time series data will be presented. The comprehensive approach, which leverages Self-organizing maps, Principal Component Analysis, and k-means, was meticulously executed to reveal insights into monthly energy usage behaviors. A key metric utilized to evaluate the effectiveness of the clustering methodology is the Silhouette Score. The Silhouette Score quantifies the separation between clusters, offering a means to gauge the quality of clustering outcomes. Ranging from -1 to 1, a higher Silhouette Score indicates well-defined clusters with considerable separation, while a lower score implies potential cluster overlap. The silhouette score vs. the number of clusters is shown in Figure 2 which indicates that 24 clusters or K= 24 is the best choice for an optimized clustering of the dataset in the K-means clustering phase. Through rigorous experimentation and analysis, the integrated SOM-k-means methodology yielded a Silhouette Score of 0.66. This score underscores the remarkable efficacy of the approach in uncovering distinctive energy consumption patterns within the time series data. The achieved score surpasses the threshold indicative of meaningful cluster separation, affirming the utility of the methodology in enhancing insights into energy usage behaviors. The obtained Silhouette Score not only validates the proficiency of the integrated methodology but also underscores its potential real-world applications. A score of 0.66 signifies a substantial degree of separation between clusters, implying that the clusters accurately represent distinct energy consumption patterns within the data. This outcome has far-reaching implications for diverse sectors, including energy management, demand forecasting, and policy formulation, even in different types of time series data other than energy consumption data recorded from smart meters. Beyond the Silhouette Score, the clustering outcomes themselves provide invaluable insights into energy consumption behaviors for further analysis of the dataset.

The integrated approach successfully distilled complex data into clusters that exhibit coherent and interpretable patterns. By leveraging the abstraction power of SOMs, the dimensionality reduction capabilities of PCA, and the refining prowess of k-means, the methodology facilitates the identification of unique energy usage trends across various temporal scales.

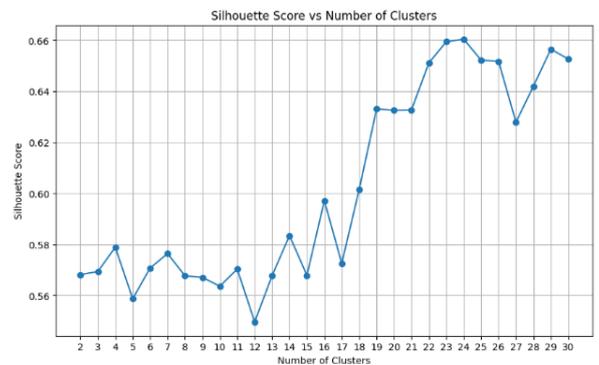

Figure 2.  Silhouette score of the K-means algorithm resulting in an optimal number of 24 clusters and a silhouette score of 0.66

## V. CONCLUSION

This paper introduced a novel and comprehensive methodology that synergistically combines Self-organizing maps and k-means for the purpose of uncovering intricate energy consumption patterns within time series data. The integration of these techniques demonstrated a powerful approach to extracting meaningful insights from complex energy consumption behaviors. The findings of this study underscore the efficacy of the proposed methodology. With a Silhouette Score of 0.66, the clustering outcomes affirm the methodology's ability to accurately capture distinct energy consumption patterns within the data. The significance of this achievement is far-reaching, transcending the realms of energy management and policy formulation to impact areas such as demand forecasting, resource optimization, and environmental conservation. The synergistic combination of SOM and k-means proved instrumental in enhancing the accuracy, efficiency, and interpretability of energy consumption pattern analysis. The abstraction prowess of SOM facilitated the initial clustering, while PCA's dimensionality reduction capabilities streamlined the subsequent k-means clustering. This fusion resulted in clusters that represent coherent and interpretable energy usage behaviors. Furthermore, this study not only presented a robust methodology but also illuminated potential avenues for future research. Fine-tuning parameters, exploring alternative dimensionality reduction techniques and other clustering algorithms, and applying the approach to diverse datasets offer opportunities for enhancing its performance and applicability. It's important to mention that a more thorough dataset would definitely help with the results since the dataset used in this research wasn't complete. For instance, some of the houses didn't have all of the dates in each month which made it difficult to use all of the dataset so only a selected number of houses could be used in the project.

In conclusion, the integrated SOM-k-means methodology showcased its prowess in unraveling energy consumption patterns within time series data. The achieved insights are poised to empower informed decision-making, optimize energy usage strategies, and contribute to sustainable practices. As energy consumption analysis assumes increasing importance in our world, the approach introduced in this paper stands as a valuable tool that bridges the gap between complex data and actionable insights.